\documentclass[runningheads]{llncs}
\usepackage[utf8]{inputenc}
\usepackage[T1]{fontenc}
\usepackage{newtxtext}
\usepackage[varvw]{newtxmath}
\usepackage{graphicx}
\usepackage{url}
\usepackage{booktabs}
\usepackage{cite}
\usepackage{amsmath}
\usepackage{microtype}
\usepackage{algorithm}
\usepackage{algpseudocode}
\usepackage{flafter}
\usepackage{float}
\usepackage[nameinlink,noabbrev]{cleveref}

\newcommand{\pp}{\,\text{pp}}

\crefname{algorithm}{Algorithm}{Algorithms}
\crefname{figure}{Fig.}{Figs.}\Crefname{figure}{Fig.}{Figs.}
\crefname{table}{Table}{Tables}\Crefname{table}{Table}{Tables}
\crefname{section}{Sect.}{Sects.}\Crefname{section}{Section}{Sections}

\begin{document}
\flushbottom
\emergencystretch=2em

\title{Beyond Node Attributes: A Controlled Evaluation and Ablation of Local
Interaction Statistics for Ethereum Phishing Detection}
\titlerunning{Local Interaction Statistics for Ethereum Phishing Detection}

\author{Thanh Dat Tran \and Duy Loi Duong \and Duy Luu Hoang \and
Minh Huy Tran \and Vo Minh Thu Le \and Nhat Tung Le}
\authorrunning{\mbox{}}
\institute{FPT University, Ho Chi Minh City, Vietnam\\
\email{takedo03377@gmail.com}\quad\email{tanbaobadao123@gmail.com}\\
\email{luuhoang300506@gmail.com}\quad\email{minhhuyhc1502@gmail.com} \quad\email{thulvm@fpt.edu.vn}\quad\email{tungln4@fpt.edu.vn}}

\maketitle

\begin{abstract}
Methods for detecting phishing accounts on Ethereum are usually evaluated as
complete pipelines that differ in features, architecture, and training
protocol, which makes the contribution of interaction information hard to
isolate. This paper evaluates the Unlabeled Feature Extension (UFE), a block of
12 local interaction statistics computed from each account's incoming and
outgoing \texttt{call} and \texttt{transfer} edges without reading class labels.
UFE is appended to the 14 source features of HEIG-Phish and tested with six
model families: Logistic Regression, Random Forest, LightGBM, Graph Attention
Network, GraphSAGE, and Meta-IFD. Within each pair of Baseline and Plus variants, the data
partition, model configuration, random seed, decision threshold, and sampled
neighborhoods are identical; only the input feature block changes. Across 20
predetermined stratified partitions of the same 2,763 labeled externally owned
accounts, UFE increased test Macro-F1 by 1.0 to 9.1 percentage points across
the six models. Every pair improved on all 20 partitions, all
Nadeau--Bengio corrected 95\% confidence intervals lie above zero, and the
false-positive rate decreased for all six models. An exact-edge Random Forest
ablation over 10 predetermined partitions associates most of the gain with the
incoming transfer-edge \texttt{sum} attribute; once this feature is present, an incremental
contribution of the remaining 11 UFE columns was not established. The results
support reporting feature-augmented baselines alongside graph models, but they
show stability under re-partitioning of one static, transductive benchmark
rather than temporal or cross-dataset generalization.

\end{abstract}

\section{Introduction}
\label{sec:intro}

The behavior of an Ethereum address can be described in two complementary ways.
The first uses account-level attributes, such as transaction frequency or total
value received. The second uses the position of the address in a directed
interaction graph with two edge types: value transfers (\texttt{transfer}) and
smart-contract calls (\texttt{call}). Tabular classifiers exploit the first
view directly, whereas graph neural networks (GNNs) propagate information
along the second.

The reason for treating phishing as a graph problem lies in how this fraud
operates. A phishing address collects funds from many victims, so its
signature is visible in who sends value to it, how often, and through which
relation, rather than in any single transaction. On HEIG-Phish, the median
number of incoming edge records is 11 for phishing accounts and only 2 for
normal accounts. Part of this relational signal can be summarized as a few
per-account statistics computed directly from the typed edges. If such
statistics are informative, both tabular models and GNNs can use them
directly. Existing comparisons, however, change architecture, features, and
hyperparameters together and use one fixed partition, so they do not show how
much a common block of interaction statistics adds to each type of model.

This paper addresses that gap with a controlled design. We define UFE as 12
local interaction statistics that count and sum the incoming and outgoing
\texttt{call} and \texttt{transfer} edges of each account without reading
class labels. We study six model families: Logistic Regression (LR), Random
Forest (RF), LightGBM, Graph Attention Network (GAT), GraphSAGE, and Meta-IFD.
For each family, the \emph{Baseline} variant receives only the 14 account
features distributed with HEIG-Phish (the source features), and the
\emph{Plus} variant receives these inputs with UFE appended, under an identical partition, seed,
configuration, sampled data, and threshold. The study has two aims: to measure
whether adding the same label-free block improves each model across 20
predetermined partitions, and to identify which part of the block produces the
improvement, in particular the summed incoming transfer-edge attribute
(\texttt{trans\_in\_sum}) relative to the remaining statistics.

The contributions are threefold: an explicit, label-free construction of the
12 statistics (\cref{alg:ufe}); an estimate of the matched effect across six
models and 20 predetermined partitions using corrected paired inference with
multiplicity control, reporting every pair regardless of outcome; and an
exact-edge ablation with a shuffled-feature control that identifies the
dominant statistic. A post-hoc audit further shows that the released source
features cannot be reconstructed from the released edge tables.

\section{Related Work}
\label{sec:related}

Early work on illicit-account detection treated blockchain activity as a
source of account-level behavioral features. Farrugia et al. derived
transaction-based account attributes and trained an XGBoost classifier to
separate illicit from licit Ethereum accounts~\cite{farrugia2020}. For
phishing, Chen et al. combined cascade feature extraction on a transaction
graph with a LightGBM ensemble designed for class imbalance
\cite{chen2020phishing}. Such compact statistics can support accurate
classification without an end-to-end GNN.

Graph-based methods retain relations that tabular summaries may lose, for
example \textit{trans2vec} with transaction amount and time
\cite{wu2022phishers}, graph convolution combined with an autoencoder
\cite{chen2021gcn}, and multi-edge feature graphs with bagging
\cite{zhang2023phishing}, built on backbones such as
GraphSAGE~\cite{hamilton2017sage} and GAT~\cite{velickovic2018gat}. TTAGN
places temporal transaction records on edges~\cite{li2022ttagn}, FA-GNN
filters noisy neighbors~\cite{liu2022fagnn}, and dynamic feature fusion
combines global graph representations with local transaction semantics
\cite{zhang2025dynamic}. At the heterogeneous-graph level, HAN applies
node-level and semantic-level attention~\cite{wang2019han}, HGT uses
type-specific transformations for nodes and edges~\cite{hu2020hgt}, and
Meta-IFD models fine-grained meta-interactions among externally owned accounts
(EOAs), contract accounts (CAs), transfers, and contract calls, using an
interaction-aware conditional variational autoencoder to generate additional
feature views~\cite{jin2025}. Mean and attention aggregators, as in GraphSAGE
and GAT, cannot count how many neighbors a node has, whereas sum aggregation
can~\cite{xu2019gin}, and simple models with well-chosen features or label
propagation often match GNNs~\cite{ivanov2021bgnn,huang2021cs}. Because these complete methods change several
components at once, comparisons among them cannot isolate the value of one
block of interaction statistics shared across different backbones.

For evaluation, common tests have inflated Type~I error when training and test
sets overlap across repetitions~\cite{dietterich1998approximate}, which
motivates corrected resampled inference~\cite{nadeau2003inference}, Holm
correction~\cite{holm1979simple}, and equivalence
testing~\cite{walker2011equivalence}.

\section{Data and Method}
\label{sec:methods}

\begin{figure}[H]
\centering
\includegraphics[width=\textwidth]{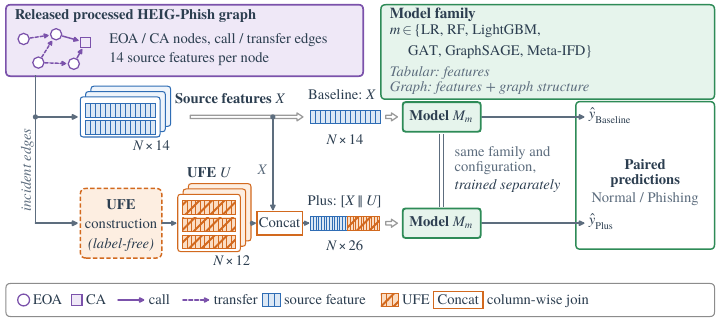}
\caption{Overview of the UFE evaluation framework.}
\label{fig:framework}
\end{figure}

\Cref{fig:framework} summarizes the evaluation framework. The
processed graph provides the 14 source features and the typed incoming and
outgoing edges of each account, from which UFE is computed without labels.
Baseline receives $X$ and Plus receives $[X\Vert U]$; the two variants use the
same model family and configuration but are trained separately, and graph
models receive the same graph structure in both.

\subsection{HEIG-Phish Data}
\label{sec:data}

We use the processed phishing dataset released with Meta-IFD, modelled as a
heterogeneous Ethereum interaction graph (HEIG)~\cite{jin2025,metaifdcode}; we
call it HEIG-Phish. The graph contains EOAs and CAs connected
by two directed relations, \texttt{call} and \texttt{trans} (transfer).
\Cref{tab:dataset} lists node, label, and edge counts; the processed file is
pinned by SHA-256 \texttt{8cc653bb\ldots f690e8}. Each edge carries two numerical attributes whose published column names
are \texttt{count} and \texttt{sum}; because their construction is not
documented, we use them only under these names and denote them $a_0$ and $a_1$.
The raw tables contain no timestamp or block-number field.

\begin{table}[H]
\centering
\small
\caption{HEIG-Phish as used in this study (released processed graph).}
\label{tab:dataset}
\begin{tabular}{lr@{\qquad}lr}
\toprule
\multicolumn{2}{l}{Nodes and labels} & \multicolumn{2}{l}{Directed edge records} \\
\midrule
EOA nodes & 22,429,276 & EOA $\xrightarrow{\text{trans}}$ EOA & 26,293,882 \\
CA nodes & 2,279,349 & EOA $\xrightarrow{\text{call}}$ CA & 20,425,382 \\
Labeled EOAs & 2,763 & EOA $\xrightarrow{\text{trans}}$ CA & 2,994,136 \\
\quad phishing / normal & 1,206 / 1,557 & CA $\xrightarrow{\text{trans}}$ EOA & 2,096,037 \\
Released split (train/val/test) & 1,657 / 553 / 553 & CA $\xrightarrow{\text{call}}$ CA & 1,879,385 \\
Source features / UFE & 14 / 12 & CA $\xrightarrow{\text{trans}}$ CA & 527,653 \\
 & & Total & 54,216,475 \\
\bottomrule
\end{tabular}
\end{table}

The task is binary classification of the 2,763 labeled EOAs: 1,206 phishing and
1,557 normal. These nodes correspond to 2,762 distinct addresses, because one
address appears twice in the label list with conflicting labels; we keep the
released labels and nodes unchanged for comparability with the source benchmark
and discuss the consequence in \cref{sec:discussion}. Each node has the 14
account features released with the benchmark, whose construction code is not
included in the release. The main experiments use the full static graph for
label-free neighborhood materialization, so the setting is transductive: the
graph structure around test nodes is present when features are produced, but
test labels are used neither to construct UFE nor to train models.

\subsection{Constructing UFE from the Graph}
\label{sec:ufe}

\begin{figure}[H]
\centering
\includegraphics[width=0.90\textwidth]{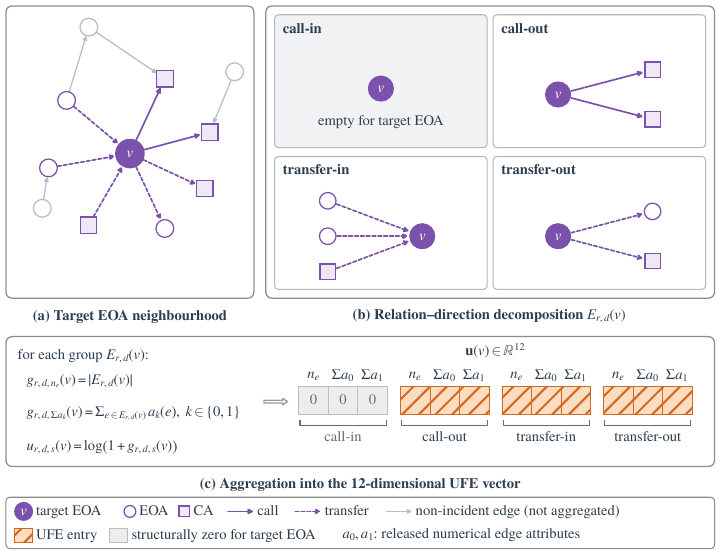}
\caption{Construction of the 12 UFE statistics for a target EOA $v$.}
\label{fig:ufe}
\end{figure}

For a target node $v$, relation $r\in\{\texttt{call},\texttt{trans}\}$, and
direction $d\in\{\mathrm{in},\mathrm{out}\}$, let $E_{r,d}(v)$ be the set of
edges of relation $r$ that enter $v$ ($d=\mathrm{in}$) or leave $v$
($d=\mathrm{out}$). For each set we compute three statistics
$s\in\{n_e,\Sigma a_0,\Sigma a_1\}$:
\begin{equation}
g_{r,d,n_e}(v)=|E_{r,d}(v)|,\qquad
g_{r,d,\Sigma a_k}(v)=\sum_{e\in E_{r,d}(v)} a_k(e),\quad k\in\{0,1\},
\end{equation}
and the feature supplied to the model is
\begin{equation}
u_{r,d,s}(v)=\log\!\left(1+g_{r,d,s}(v)\right).
\end{equation}
Two relations, two directions, and three statistics give $2\times2\times3=12$
values (\cref{alg:ufe}). \Cref{fig:ufe} illustrates the three steps: (a) only
edges incident to $v$ are used, (b) they are split into four relation--direction
groups, and (c) each group yields $n_e$, $\Sigma a_0$, and $\Sigma a_1$. The $n_e$ statistic counts processed
edge records and does not assume that one record corresponds to one raw
transaction. The released schema contains no \texttt{call} edge ending at an
EOA, so the three call-in statistics are always zero for the labeled targets
and only nine columns vary. The logarithm compresses long-tailed values; it
changes the scale seen by linear and neural models but not the ordering used by
trees. No step reads a class label.

UFE is computed in two ways. In the six-model evaluation, it is computed inside
two-layer sampled neighborhoods with fanout $[100,100]$ per edge type and batch
size 512, and the two variants of a pair receive exactly the same sampled data.
In the ablation, the statistics are computed deterministically from all
relevant edges; we call these \emph{exact-edge} statistics.

\begin{algorithm}[!b]
\caption{Construction of Local Interaction Features (UFE)}
\label{alg:ufe}
\footnotesize
\begin{algorithmic}[1]
\Require Directed heterogeneous graph with typed edges $e=(s_e,t_e)$, relation
$r(e)\in\{\texttt{call},\texttt{trans}\}$, released attributes
$(a_{0,e},a_{1,e})=(\texttt{count},\texttt{sum})$; target nodes $T$; edge set
$E'$ (all edges for exact UFE, or the edges of the sampled two-hop batch for
sampled UFE)
\Ensure $U\in\mathbb{R}^{|T|\times 12}$
\For{each $v\in T$}
  \For{each $r\in\{\texttt{call},\texttt{trans}\}$ and $d\in\{\mathrm{in},\mathrm{out}\}$}
    \State $E_{r,d}(v)\gets\{e\in E' : r(e)=r,\ t_e=v \text{ if } d=\mathrm{in},\ s_e=v \text{ if } d=\mathrm{out}\}$
    \State $n_e\gets|E_{r,d}(v)|$;\quad $\alpha_0\gets\sum_{e\in E_{r,d}(v)} a_{0,e}$;\quad $\alpha_1\gets\sum_{e\in E_{r,d}(v)} a_{1,e}$
    \For{each $x\in\{n_e,\alpha_0,\alpha_1\}$}
      \State append $\log\!\big(1+\max(0,\mathrm{nan\_to\_num}(x))\big)$ to $U_v$
    \EndFor
  \EndFor
\EndFor
\State \Return $U$ \Comment{no class label is read}
\end{algorithmic}
\end{algorithm}

\subsection{Matched Comparison Design and Evaluation Protocol}
\label{sec:protocol}

Every Plus variant receives 26 inputs: the 14 source features and the 12 UFE
statistics. Within each pair, the partition, sampled data, model seed,
configuration, threshold, and validation checkpoint rule are held fixed, so the
comparison isolates the added feature block and does not define six new
algorithms. LR uses standardization (fitted on training nodes) followed by
logistic regression with $C=10$, the \texttt{liblinear} solver, and at most
5,000 iterations. RF uses 100 trees, maximum depth 10, minimum leaf size 1, and
no class weighting~\cite{breiman2001rf}. LightGBM uses a binary objective, 300
estimators, and learning rate 0.01~\cite{ke2017lightgbm}; these three models are
implemented with scikit-learn and LightGBM~\cite{pedregosa2011sklearn}. GAT
applies batch normalization to the source features, a linear node encoder, and
two GAT layers with four attention heads, hidden width 32, dropout 0.2,
learning rate 0.01, and weight decay $5\times10^{-4}$~\cite{velickovic2018gat}.
GraphSAGE uses the same settings with two mean-aggregation
layers~\cite{hamilton2017sage}. The source paper reports
search spaces rather than the final Phish configuration, so our Meta-IFD
baseline uses the defaults of the released code: hidden width 128, learning
rate 0.001, contrastive weight 0.1, three generated views, and weight decay
$5\times10^{-4}$~\cite{metaifdcode}. Meta-IFD+ keeps the frozen 14-dimensional
generator unchanged and concatenates the same 12 UFE values to every generated
view of both EOA and CA nodes, increasing each view from 14 to 26 dimensions;
the two variants share sampled batches and generated views. Neural models (PyTorch and PyTorch
Geometric~\cite{paszke2019pytorch,fey2019pyg}) train for at most 1,000 epochs
with early-stopping patience 50 on validation Macro-F1; the threshold is 0.5 and
model seeds are 12--16.

The evaluation proceeds in three steps. The first uses the released 60/20/20
split (1,657/553/553 nodes) to check how closely the reported baselines are
reproduced; this test partition was seen during development. The second is the main
analysis over 20 partitions with partition seeds 2001--2020, each a stratified
random 60/20/20 division of the same 2,763 labeled EOAs. Configurations were
fixed in advance and not tuned further; every pair and partition is reported,
and test data were materialized only after the checkpoint was frozen. Within a
partition, Macro-F1 is averaged over five model seeds, and the paired
difference $d_i=F^{+}_i-F^{\mathrm{base}}_i$ is the unit of analysis. Because
partitions overlap, we use the corrected resampled $t$ procedure of Nadeau and
Bengio~\cite{nadeau2003inference}:
\begin{equation}
\widehat{\mathrm{SE}}_{\mathrm{NB}}
=\sqrt{\left(\frac{1}{k}+\frac{n_{\mathrm{test}}}{n_{\mathrm{train}}}\right)s_d^2},
\end{equation}
where $k=20$, $s_d^2$ is the sample variance of the paired differences, and
$n_{\mathrm{test}}/n_{\mathrm{train}}=553/1657$. We report two-sided 95\%
intervals with 19 degrees of freedom and Holm adjustment across the six
pairs~\cite{holm1979simple}. The predefined success rule requires a mean
improvement of at least $0.5\pp$, a lower confidence bound above zero, and a
Holm-adjusted $p<0.05$.

The third step is an exact-edge RF ablation on 10 predefined stratified
partitions (seeds $101,202,\ldots,1010$) with five model seeds each. The feature
configurations are listed in \cref{fig:ablation}(a); the shuffled control permutes
UFE rows within each partition, which breaks the association between nodes and
features but keeps the marginal distribution. Within
each partition and configuration, the number of trees $\{100,200,300\}$ and
maximum depth $\{10,\texttt{None}\}$ are selected on validation Macro-F1. The four
superiority comparisons in \cref{fig:ablation}(b) form one Holm family, and each
must meet the success rule with $k=10$. The permuted control is assessed by
corrected two one-sided tests (TOST) with an equivalence margin of
$\pm0.5\pp$~\cite{walker2011equivalence}. A supplementary analysis, specified
after the primary one, repeats this design on the 20 main partitions as a
robustness check.

The primary metric is Macro-F1; we also report positive-class precision,
recall, and F1, Micro-F1 (as reported by the source paper), and the
false-positive rate (FPR)~\cite{saito2015precision}. Statistical decisions use
partition-level paired differences, not model seeds, as observations.

\section{Results}
\label{sec:results}

\subsection{Baseline Reproduction and the Effect of UFE on Six Models}
\label{sec:res-main}

\Cref{tab:reproduction} compares the values
reported in the source paper with our reproduced baselines on that
split; LR and RF are not reported there, and an SD of 0.00 means that all seeds
gave identical predictions. Full Baseline--Plus results on this development-used
split are reported in the supplementary material rather than used as
confirmatory evidence. LightGBM and Meta-IFD are reproduced within $0.3\pp$ of the
reported Macro-F1, whereas GraphSAGE and GAT are lower by $1.37\pp$ and
$5.96\pp$; the GAT baseline is therefore not a numerical reproduction of the
reported result.

\begin{table}[H]
\centering
\caption{Reported and reproduced baselines on the released HEIG-Phish split
(\%, mean $\pm$ SD over five model seeds; column-best values in bold).}
\label{tab:reproduction}
\renewcommand{\arraystretch}{1,05}
\setlength{\tabcolsep}{3pt}   
\resizebox{\textwidth}{!}{%
\begin{tabular}{llcccccr}
\toprule
Model & Source & Precision & Recall & Binary-F1 & Micro-F1 & Macro-F1 & $\Delta$ Macro \\
\midrule
LightGBM & Paper & 94.13$\pm$0.67 & 97.01$\pm$1.13 & 95.54$\pm$0.37 & 96.06$\pm$0.31 & 96.00$\pm$0.32 &  \\
 & Reproduced & 93.65$\pm$0.00 & 97.93$\pm$0.00 & 95.74$\pm$0.00 & 96.20$\pm$0.00 & 96.16$\pm$0.00 & $+0.16$ \\
\midrule
GAT & Paper & 90.20$\pm$1.52 & 95.19$\pm$0.86 & 92.62$\pm$0.95 & 93.38$\pm$0.90 & 93.31$\pm$0.90 &  \\
 & Reproduced & 81.51$\pm$3.56 & 92.12$\pm$2.99 & 86.47$\pm$2.87 & 87.41$\pm$2.76 & 87.35$\pm$2.76 & $-5.96$ \\
\midrule
GraphSAGE & Paper & 93.46$\pm$0.45 & 96.02$\pm$0.62 & 94.72$\pm$0.44 & 95.33$\pm$0.39 & 95.27$\pm$0.39 &  \\
 & Reproduced & 90.68$\pm$0.97 & 96.02$\pm$1.08 & 93.27$\pm$0.79 & 93.96$\pm$0.71 & 93.90$\pm$0.72 & $-1.37$ \\
\midrule
Meta-IFD & Paper & \textbf{97.37$\pm$0.55} & \textbf{98.01$\pm$0.17} & \textbf{97.68$\pm$0.27} & \textbf{97.97$\pm$0.24} & \textbf{97.94$\pm$0.24} &  \\
 & Reproduced & 97.19$\pm$0.71 & 97.51$\pm$0.59 & 97.35$\pm$0.22 & 97.69$\pm$0.20 & 97.65$\pm$0.20 & $-0.29$ \\
\bottomrule
\end{tabular}}
\end{table}

\begin{figure}[t]
\centering
\includegraphics[width=0.90\textwidth]{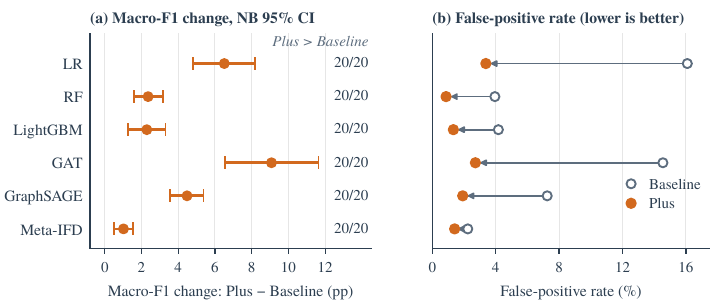}
\caption{Matched effect of UFE over 20 predetermined partitions.}
\label{fig:matched}
\end{figure}

Across the 20 predetermined partitions, every Plus variant has higher mean
Macro-F1 than its matched baseline (\cref{tab:matched,fig:matched}); $\Delta$ in
\cref{tab:matched} is the paired Macro-F1 difference with its NB 95\% interval,
and the FPR values in \cref{fig:matched}(b) are descriptive means without
intervals. Gains range
from $1.02\pp$ for Meta-IFD+ to $9.08\pp$ for GAT+, and each Plus variant is
better on all 20 partitions. All six corrected intervals lie above zero, all
Holm-adjusted $p$ values are below .001, and every pair meets the predefined
success rule. FPR decreases for every model, most strongly for LR
($16.09\%\rightarrow3.38\%$) and GAT ($14.55\%\rightarrow2.72\%$), while the
recall of every Plus variant remains above $96.9\%$. The improvement therefore
comes mainly from fewer false positives rather than from higher recall alone.

\begin{table}[tbp]
\centering
\caption{Matched results across 20 predetermined HEIG-Phish partitions
(\%, mean $\pm$ SD; column-best values in bold and lower FPR is better).}
\label{tab:matched}
\renewcommand{\arraystretch}{1,05}
\setlength{\tabcolsep}{3pt}     
\resizebox{\textwidth}{!}{%
\begin{tabular}{llccccccc}
\toprule
Model & Variant & Precision & Recall & Binary-F1 & Macro-F1 & FPR & $\Delta$ Macro-F1 & NB 95\% CI \\
\midrule
LR & Baseline & 82.59$\pm$2.09 & \textbf{98.51$\pm$0.92} & 89.84$\pm$1.35 & 90.25$\pm$1.40 & 16.09$\pm$2.35 &  &  \\
 & Plus & 95.69$\pm$0.88 & 97.05$\pm$1.36 & 96.36$\pm$0.76 & 96.76$\pm$0.66 & 3.38$\pm$0.73 & +6.51 & [4.81, 8.20] \\
\midrule
RF & Baseline & 94.94$\pm$1.21 & 95.60$\pm$1.67 & 95.26$\pm$0.96 & 95.79$\pm$0.84 & 3.95$\pm$0.99 &  &  \\
 & Plus & \textbf{98.85$\pm$0.47} & 96.96$\pm$1.17 & 97.89$\pm$0.62 & 98.15$\pm$0.53 & \textbf{0.87$\pm$0.36} & +2.36 & [1.57, 3.15] \\
\midrule
LightGBM & Baseline & 94.70$\pm$1.26 & 95.87$\pm$1.71 & 95.27$\pm$0.83 & 95.79$\pm$0.73 & 4.17$\pm$1.06 &  &  \\
 & Plus & 98.26$\pm$0.64 & 97.39$\pm$1.37 & 97.82$\pm$0.82 & 98.08$\pm$0.72 & 1.33$\pm$0.49 & +2.29 & [1.28, 3.31] \\
\midrule
GAT & Baseline & 83.09$\pm$2.63 & 91.59$\pm$2.93 & 87.06$\pm$2.09 & 88.03$\pm$1.94 & 14.55$\pm$2.68 &  &  \\
 & Plus & 96.52$\pm$0.97 & 97.01$\pm$1.30 & 96.75$\pm$0.68 & 97.12$\pm$0.59 & 2.72$\pm$0.80 & +9.08 & [6.53, 11.64] \\
\midrule
GraphSAGE & Baseline & 91.06$\pm$1.96 & 94.57$\pm$1.88 & 92.74$\pm$1.03 & 93.46$\pm$0.93 & 7.25$\pm$1.77 &  &  \\
 & Plus & 97.54$\pm$0.98 & 97.84$\pm$1.07 & 97.68$\pm$0.59 & 97.94$\pm$0.52 & 1.92$\pm$0.80 & +4.48 & [3.57, 5.38] \\
\midrule
Meta-IFD & Baseline & 97.11$\pm$0.97 & 96.91$\pm$1.13 & 97.00$\pm$0.53 & 97.34$\pm$0.47 & 2.24$\pm$0.79 &  &  \\
 & Plus & 98.19$\pm$0.68 & 98.12$\pm$0.81 & \textbf{98.15$\pm$0.49} & \textbf{98.36$\pm$0.43} & 1.41$\pm$0.54 & +1.02 & [0.50, 1.53] \\
\bottomrule
\end{tabular}}
\end{table}

\subsection{Exact-Edge Ablation with Random Forest}
\label{sec:res-ablation}

\begin{figure}[H]
\centering
\includegraphics[width=0.90\textwidth]{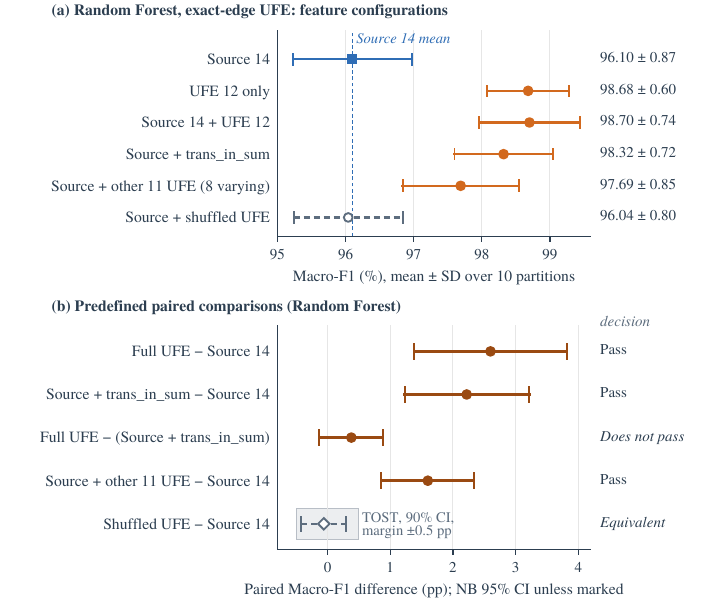}
\caption{Exact-edge Random Forest ablation over 10 predetermined partitions.}
\label{fig:ablation}
\end{figure}

In the exact-edge RF ablation, the full UFE block raises Macro-F1 from
$96.10\%$ to $98.70\%$ (\cref{fig:ablation}). Panel (a) is descriptive;
only the predefined comparisons in panel (b) support statistical conclusions. Adding only
\texttt{trans\_in\_sum} to the source features yields $98.32\%$, i.e.\
$+2.22\pp$ of the $+2.60\pp$ full-block gain, whereas adding the other 11
columns without \texttt{trans\_in\_sum} yields $97.69\%$. UFE alone reaches
$98.68\%$, close to the combined input; this figure is descriptive and not part
of the Holm family, so the two configurations cannot be declared equivalent.
Three of the four predefined comparisons meet the success rule, each with
adjusted $p=.0027$ and wins on 10/10 partitions. The remaining comparison, the
incremental contribution of the other 11 columns after \texttt{trans\_in\_sum},
does not: its mean gain is $0.38\pp$, its interval $[-0.13,0.88]$ includes zero,
and its adjusted $p=.1239$. The shuffled control is equivalent to the source
features within $\pm0.5\pp$ ($-0.06\pp$, 90\% CI $[-0.42,0.30]$, TOST
$p=.0251$). This supports the view that the gain depends on the statistics
belonging to the correct account, not on their marginal distribution.

\section{Discussion}
\label{sec:discussion}

The central result is that the same compact, label-free summary improves six
different models on every predetermined partition, including the already
strong Meta-IFD. The size of the improvement depends on the model: it is
largest where the baseline representation exposes the relevant signal least
directly and smallest for Meta-IFD, whose heterogeneous multi-view design may
already encode part of this information. The ablation narrows the picture: most
of the RF gain is associated with \texttt{trans\_in\_sum}, and an additional
contribution of the other statistics was not established. Because the ablation
uses exact aggregation while the main evaluation uses sampled neighborhoods, it
characterizes the added information but does not decompose the six-model gains
directly. GAT and GraphSAGE gain the most, which is consistent with their mean
and attention aggregators being unable to recover neighbor counts that UFE
supplies explicitly~\cite{xu2019gin}. Because the reproduced GAT is $5.96\pp$ below its reported value, its
large gain should be read as an upper estimate.

\textbf{Post-hoc benchmark audit.} The dominant statistic resembles one of the source features, so we examined the
relation between the two blocks in post-hoc, descriptive analyses
(Supplementary Fig.~S1). Of the 168 pairs of one source feature and one
exact-edge UFE column, 108 contain no constant column; no pair reaches a $95\%$
near-duplicate threshold, and after excluding observations where both values
are zero, the highest fraction of equal or close values is only $58.95\%$. The
strongest predictive pair, \texttt{trans\_in\_sum} and \texttt{V\_inv\_t}, has
a logarithmic Pearson correlation of $r=0.905$ but agrees for only $45.5\%$ of
accounts, with lower agreement for phishing than for normal accounts. Testing
252 candidate formulas per source feature (edge counts, sums of \texttt{count}
and \texttt{sum}, ratios, and differences) also failed to reconstruct any
non-constant source feature for at least $99\%$ of accounts. The 14 source
features are thus not derivable from the edge tables, and UFE is not a
duplicate of them. Reusing the frozen RF predictions of the 20-partition
analysis, the UFE gain was $+0.31\pp$ among accounts whose two values agree and
$+4.70\pp$ among accounts whose values differ, although the agreeing group was
already close to ceiling. The gain may therefore reflect information that the
released features express differently, a benchmark-specific effect, or both;
these analyses are post hoc and not confirmatory.

In practice, explicit interaction summaries are a simple way to reduce false
positives across very different models; we did not compare computational cost.

Regarding internal validity, model selection used validation data only, the
largest mean gap between training and test Macro-F1 was $3.2\pp$, and no GNN met
the predefined criterion for harmful over-smoothing.

The scope of the study remains limited. All partitions reuse the same 2,763
labeled accounts, so the evidence concerns stability within one benchmark
rather than generalization to a new collection period, label source, or account
population. The graph is static and transductive, and the release provides no
transaction timestamps. If the edge records extend beyond the period used to
compute the source features or beyond the time at which phishing labels were
assigned, UFE may encode post-label activity correlated with the class; the
concentration of gains among accounts whose edge-derived and released values
disagree is consistent with this possibility. We therefore cannot rule out that
part of the gain reflects temporal leakage in the benchmark rather than
information available at detection time. The source
paper names the platforms from which labels were collected but not how normal
accounts were selected; \texttt{trans\_in\_sum} alone separates the two classes
with an AUC of $0.968$ (post hoc), so the benchmark may be comparatively easy
and sampling artifacts cannot be excluded. The duplicated address with
conflicting labels has one copy in training and one in test in the released
split, affecting one of 553 test nodes. Finally, the GAT and GraphSAGE
baselines are matched controls, not exact reproductions of the published
models.

\section{Conclusion}
\label{sec:conclusion}

This study evaluated a block of 12 label-free local interaction statistics for
Ethereum phishing detection under matched conditions. On HEIG-Phish, the same
block improved linear, tree-based, graph, and self-supervised heterogeneous
models on every predetermined partition, mainly by reducing false positives,
and most of the gain was associated with \texttt{trans\_in\_sum}. Because the
released source features cannot be reconstructed from the edge records and the
benchmark has no timestamps, the results are conditional on how the benchmark
was constructed. Future work should apply the same frozen feature definition to
a second fraud type with contract-account targets, such as HEIG-Ponzi, and to
independently collected, time-ordered data, and should compute exact UFE for the
tabular models and add a sum-aggregation GNN baseline such as GIN.

\subsubsection*{Data, Code, and Ethics.}
The original Meta-IFD implementation is publicly available~\cite{metaifdcode};
the authors' notebooks, frozen protocols, and result tables are available in
the anonymized review repository at
\url{https://anonymous.4open.science/r/ufe-review-materials}.
The models produce research risk scores, not findings of wrongdoing, so any
practical use requires human review.

\bibliographystyle{splncs04}
\bibliography{references}

\end{document}